\documentclass{article}
\usepackage{ijcai19}

\usepackage{times}
\usepackage{soul}
\usepackage[hyphens]{url}
\usepackage[breaklinks,dvipdfmx,hidelinks]{hyperref}
\usepackage[utf8]{inputenc}
\usepackage[small]{caption}
\usepackage[dvipdfmx]{graphicx}
\usepackage{amsmath}
\usepackage{booktabs}
\usepackage{amssymb}
\usepackage{bm}
\usepackage{cite}
\usepackage{cleveref}
\usepackage{latexsym}
\usepackage{multirow}
\usepackage{pifont}
\usepackage{siunitx}

\newcommand{\secref}[1]{\mbox{Section~\ref{#1}}}
\newcommand{\figref}[1]{\mbox{Figure~\ref{#1}}}
\newcommand{\tabref}[1]{\mbox{Table~\ref{#1}}}
\newcommand{\eqnref}[1]{\mbox{Equation~\ref{#1}}}

\makeatletter
    \let\@internalcite\cite
    \def\cite{\def\citeauthoryear##1##2{##1, ##2}\@internalcite}
    \def\shortcite{\def\citeauthoryear##1{##2}\@internalcite}
    \def\@biblabel#1{\def\citeauthoryear##1##2{##1, ##2}[#1]\hfill}
\makeatother

\crefformat{footnote}{#2\footnotemark[#1]#3}

\title{Robust Audio Adversarial Example for a Physical Attack}

\author{
Hiromu Yakura$^{1,2}$\footnote{Contact Author}\And
Jun Sakuma$^{1,2}$\\
\affiliations
$^1$University of Tsukuba \\
$^2$RIKEN Center for Advanced Intelligence Project \\
\emails
hiromu@mdl.cs.tsukuba.ac.jp,
jun@cs.tsukuba.ac.jp
}

\begin{document}

\maketitle

\begin{abstract}
We propose a method to generate audio adversarial examples that can attack a state-of-the-art speech recognition model in the physical world.
Previous work assumes that generated adversarial examples are directly fed to the recognition model, and is not able to perform such a physical attack because of reverberation and noise from playback environments.
In contrast, our method obtains robust adversarial examples by simulating transformations caused by playback or recording in the physical world and incorporating the transformations into the generation process.
Evaluation and a listening experiment demonstrated that our adversarial examples are able to attack without being noticed by humans.
This result suggests that audio adversarial examples generated by the proposed method may become a real threat.
\end{abstract}

\section{Introduction}
\label{sec:introduction}

In recent years, deep learning has achieved vastly improved accuracy, especially in fields such as image classification and speech recognition, and has come to be used practically \cite{DBLP:journals/nature/LeCunBH15}.
On the other hand, deep learning methods are known to be vulnerable to adversarial examples \cite{DBLP:journals/corr/SzegedyZSBEGF13,DBLP:journals/corr/GoodfellowSS14}.
More specifically, an attacker can make deep learning models misclassify examples by intentionally adding a small perturbation to the examples.
Such examples are referred to as adversarial examples.

\begin{figure*}[tb]
    \begin{center}
        \includegraphics[width=0.96\textwidth]{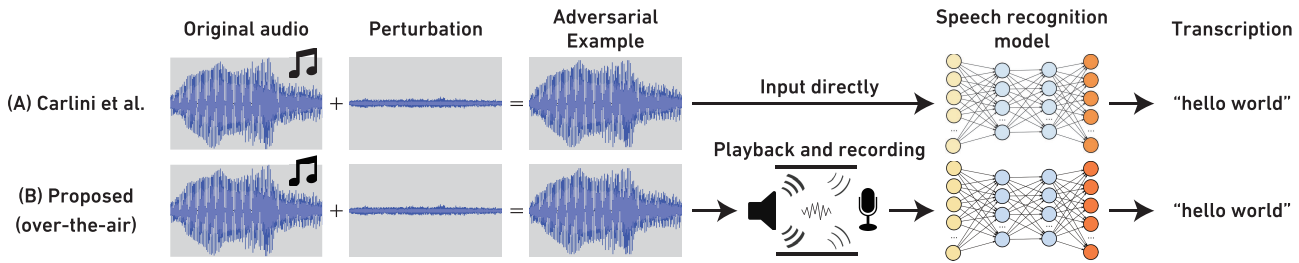}
        \caption{Illustration of the proposed attack. \protect\cite{DBLP:conf/sp/Carlini018} assumed that adversarial examples are provided directly to the recognition model. We propose a method that targets an over-the-air condition, which leads to a real threat.}
        \label{fig:overview}
    \end{center}
\end{figure*}

While many papers discussed image adversarial examples against image classification models, little research has been done on audio adversarial examples against speech recognition models,
even though speech recognition models are widely used at present in commercial applications like Amazon Alexa, Apple Siri, Google Assistant, and Microsoft Cortana and devices like Amazon Echo and Google Home.
For example, \cite{DBLP:conf/sp/Carlini018} proposed a method to generate audio adversarial examples against DeepSpeech \cite{DBLP:journals/corr/HannunCCCDEPSSCN14}, which is a state-of-the-art speech recognition model.
However, this method targets the case in which the waveform of the adversarial example is input directly to the model, as shown in \figref{fig:overview}(A).
In other words, it is not feasible to attack in the case that the adversarial example is played by a speaker and recorded by a microphone in the physical world (hereinafter called the \textit{over-the-air} condition), as shown in \figref{fig:overview}(B).

The difficulty of such an over-the-air attack can be attributed to the reverberation of the environment and noise from both the speaker and the microphone.
More specifically, in the case of the direct input, adversarial examples can be generated by determining a single data point that fools the targeted model using an optimization algorithm for a clearly described objective.
In contrast, under the over-the-air condition, adversarial examples are required to be robust against unknown environments and equipment.

Considering that audio signals spread through the air, the impact of a physical attack using audio adversarial examples would be larger than that using image adversarial examples.
For an attack scenario using an image adversarial example, the adversarial example must be presented explicitly in front of an image sensor of the attack target, e.g., the camera of an auto-driving car.
In contrast, audio adversarial examples can simultaneously attack numerous targets by spreading via outdoor speakers or radios.
If an attacker hijacks the broadcast equipment of a business complex, it will be possible to attack all the smartphones owned by people inside via a single playback of the audio adversarial example.

In the present paper, we propose a method by which to generate a robust audio adversarial example that can attack speech recognition models in the physical world.
To the best of our knowledge, this is the first approach to succeed in generating such adversarial examples that can attack complex speech recognition models based on recurrent networks, such as DeepSpeech, over the air.
Moreover, we believe that our research will contribute to improving the robustness of speech recognition models by training models to discriminate adversarial examples
through a process similar to adversarial training in the image domain \cite{DBLP:journals/corr/GoodfellowSS14}.

\subsection{Related Research}
\label{sec:introduction-related}

Some studies have proposed methods to generate audio adversarial examples against speech recognition models \cite{DBLP:journals/corr/abs-1801-00554,DBLP:journals/corr/abs-1805-07820,DBLP:conf/nips/CisseANK17,DBLP:journals/corr/abs-1808-05665,DBLP:conf/sp/Carlini018}.
These methods are divided into two groups: black-box and white-box settings.

In the black-box setting, in which the attacker can only use the score that represents how close the input audio is to the desired phrase,
\cite{DBLP:journals/corr/abs-1801-00554} proposed a method to attack a speech command classification model \cite{DBLP:conf/interspeech/SainathP15}.
This method exploits a genetic algorithm to find an adversarial example, which is recognized as a specified command word.
Inspired by this method, \cite{DBLP:journals/corr/abs-1805-07820} proposed a method to attack DeepSpeech \cite{DBLP:journals/corr/HannunCCCDEPSSCN14} under the black-box setting by combining genetic algorithms and gradient estimation.
One limitation of their method is that the length of the phrase that the attacker can make the models recognize is restricted to two words at most, even when the obtained adversarial example is directly inputted.
\cite{DBLP:conf/nips/CisseANK17} performed an attack on Google Voice application using adversarial examples generated against DeepSpeech-2 \cite{DBLP:conf/icml/AmodeiABCCCCCCD16}.
The aim of their attack was changing recognition results to different words without being noticed by humans.
In other words, they could not make the targeted model output desired words and concluded that attacking speech recognition models so as to transcribe specified words ``seem(s) to be much more challenging.''
From these points, current methods in the black-box settings are not realistic for considering the attack scenario in the physical world.

In the white-box setting, in which the attacker can access the parameters of the targeted models,
\cite{DBLP:conf/uss/YuanCZLL0ZH0G18} proposed a method to attack Kaldi \cite{Povey_ASRU2011}, a conventional speech recognition model based on the combination of deep neural network and hidden Markov model.
\cite{DBLP:journals/corr/abs-1808-05665} extended the method such that generated adversarial examples are not noticed by humans using a hiding technique based on psychoacoustics.
Although \cite{DBLP:conf/uss/YuanCZLL0ZH0G18} succeeded in attacking over the air, their method is not applicable to speech recognition models based on recurrent networks, which are becoming more popular and highly functional.
For example, Google replaced its conventional model with a recurrent network based model in 2012\footnote{\url{https://ai.googleblog.com/2015/08/the-neural-networks-behind-google-voice.html}}.

In that respect, \cite{DBLP:conf/sp/Carlini018} proposed a white-box method to attack against DeepSpeech, a recurrent network based model.
However, as mentioned previously, this method succeeds in the case of the direct input, but not in the over-the-air condition, because of the reverberation of the environment and noise from both the speaker and the microphone.
Thus, the threat of the obtained adversarial example is limited regarding the attack scenario in the physical world.

\subsection{Contribution}
\label{sec:introduction-contribution}

The contribution of the present paper is two-fold:

\begin{itemize}
    \item We propose a method by which to generate audio adversarial examples that can attack speech recognition models based on recurrent networks under the over-the-air condition. Note that such a practical attack is not achievable using the conventional methods described in \secref{sec:introduction-related}. We addressed the problem of the reverberation and the noise in the physical world by simulating them and incorporating the simulated influence into the generation process.
    \item We show the feasibility of the practical attack using the adversarial examples generated by the proposed method in evaluation and a listening experiment. Specifically, the generated adversarial examples demonstrated a success rate of 100\% for the attack through both speakers and radio broadcasting, although no participants heard the target phrase in the listening experiment.
\end{itemize}

\section{Background}
\label{sec:background}

In this section, we briefly introduce an adversarial example and review current speech recognition models.

\subsection{Adversarial Example}
\label{sec:background-adversarial}

An adversarial example is defined as follows.
Given a trained classification model $f: \mathbb{R}^n \rightarrow \left \{ 1, 2, \cdots, k \right \}$ and an input sample $\bm{x} \in \mathbb{R}^n$,
an attacker wishes to modify $\bm{x}$ so that the model recognizes the sample as having a specified label $l \in \left \{ 1, 2, \cdots, k \right \}$ and the modification does not change the sample significantly:
\begin{equation}
    \tilde{\bm{x}} \in \mathbb{R}^n\, s.t.\, f \left( \tilde{\bm{x}} \right) = l \wedge \left \lVert \bm{x} - \tilde{\bm{x}} \right \rVert \leq \delta
    \label{eq:ae-definition}
\end{equation}
Here, $\delta$ is a parameter that limits the magnitude of perturbation added to the input sample and is introduced so that humans cannot notice the difference between a legitimate input sample and an input sample modified by an attacker.

Let $\bm{v} = \tilde{\bm{x}} - \bm{x}$ be the perturbation.
Then, adversarial examples that satisfy \eqnref{eq:ae-definition} can be found by optimizing this in which $\mathop{Loss}_{f}$ is a loss function that represents how distant the input data are from the given label under the model $f$:
\begin{equation}
    \mathop{\rm argmin}_{\bm{v}} \mathop{Loss}\limits_{f} \left( \bm{x} + \bm{v}, l \right) + \epsilon \left \lVert \bm{v} \right \rVert
    \label{eq:ae-derivation}
\end{equation}
By solving the problem using optimization algorithms, the attacker can obtain an adversarial example.
In particular, when $f$ is a differentiable model, such as a regular neural network, and a gradient on $\bm{v}$ can be calculated, a gradient method such as Adam \cite{DBLP:journals/corr/KingmaB14} is often used.

\subsection{Image Adversarial Example for a Physical Attack}
\label{sec:background-image}

Considering attacks on physical recognition devices (e.g., object recognition of auto-driving cars), adversarial examples are given to the model through sensors.
In the example of the auto-driving car, image adversarial examples are given to the model after being printed on physical materials and photographed by a car-mounted camera.
Through such a process, the adversarial examples are transformed and exposed to noise.
However, adversarial examples generated by \eqnref{eq:ae-derivation} are assumed to be given directly to the model and do not work for such scenarios.

In order to address this problem, \cite{DBLP:conf/icml/AthalyeEIK18} proposed a method to simulate transformations caused by printing or taking a picture and incorporate the transformations into the generation process of image adversarial examples.
This method can be represented as follows using a set of transformations $\mathcal{T}$ consisting of, e.g., enlargement, reduction, rotation, change in brightness, and addition of noise:
\begin{eqnarray}
    & \mathop{\rm argmin}\limits_{\bm{v}} \mathbb{E}_{t \sim \mathcal{T}} \! \! \! \! \! & \left[ \mathop{Loss}\limits_{f} \left( \mathop{t} \left( \bm{x} + \bm{v} \right) \! , l \right) \, \, \right. \nonumber \\
    &                                                                                    & \left. \, \, \, \, + \, \, \epsilon \left \lVert t \left( \bm{x} \right) - t \left( \bm{x} + \bm{v} \right) \right \rVert \right]
    \label{eq:ae-robust}
\end{eqnarray}
As a result, adversarial examples are generated so that images work even after being printed and photographed.

\subsection{Audio Adversarial Example}
\label{sec:background-audio}

As explained in \secref{sec:introduction-related}, \cite{DBLP:conf/sp/Carlini018} succeeded to attack against DeepSpeech, a recurrent network based model.
Here, the targeted model has time-dependency and the same approach as image adversarial examples is not applicable.
Thus, based on the fact that the targeted model uses Mel-Frequency Cepstrum Coefficient (MFCC) for the feature extraction, they implemented MFCC calculation in a differentiable manner and optimized an entire waveform using Adam \cite{DBLP:journals/corr/KingmaB14}.

In detail, the perturbation $\bm{v}$ is obtained against the input sample $\bm{x}$ and the target phrase $\bm{l}$ using the loss function of DeepSpeech as follows:
\begin{equation}
    \mathop{\rm argmin}\limits_{\bm{v}} \mathop{Loss}\limits_{f} \left( \mathop{MFCC} \left( \bm{x} + \bm{v} \right) \! , \bm{l} \right) + \epsilon \left \lVert \bm{v} \right \rVert
    \label{eq:audio-ae}
\end{equation}
Here, $\mathop{MFCC} \left( \bm{x} + \bm{v} \right)$ represents the MFCC extraction from the waveform of $\bm{x} + \bm{v}$.
They reported the success rate of the obtained adversarial examples as 100\% when inputting waveforms directly into the recognition model, but did not succeed at all under the over-the-air condition.

To the best of our knowledge, there has been no proposal to generate audio adversarial examples, which work under the over-the-air condition, targeting speech recognition models using a recurrent network.

\section{Proposed Method}
\label{sec:proposed}

In this research, we propose a method by which to generate a robust adversarial example that can attack DeepSpeech \cite{DBLP:journals/corr/HannunCCCDEPSSCN14} under the over-the-air condition.
The basic idea is to incorporate transformations caused by playback and recording into the generation process, similar to \cite{DBLP:conf/icml/AthalyeEIK18}.
We introduce three techniques: a band-pass filter, impulse response, and white Gaussian noise.

\subsection{Band-pass Filter}
\label{sec:proposed-bpf}

Since the audible range of humans is 20 to 20,000 $\si{\hertz}$, normal speakers are not made to play sounds outside this range.
Moreover, microphones are often made to automatically cut out all but the audible range in order to reduce noise.
Therefore, if the obtained perturbation is outside the audible range, the perturbation will be cut during playback and recording and will not function as an adversarial example.

Therefore, we introduced a band-pass filter in order to explicitly limit the frequency range of the perturbation.
Based on empirical observations, we set the band to 1,000 to 4,000 $\si{\hertz}$, which exhibited less distortion.
Here, the generation process is represented as follows based on \eqnref{eq:audio-ae}:
\begin{eqnarray}
    & \mathop{\rm argmin}\limits_{\bm{v}} \! \! \! \! \! & \mathop{Loss}\limits_{f} \left( \mathop{MFCC} \left( \tilde{\bm{x}} \right) \! , \bm{l} \right) + \epsilon \left \lVert \bm{v} \right \rVert \nonumber \\
    & \, where                            \! \! \! \! \! & \tilde{\bm{x}} = \bm{x} + \mathop{BPF}\limits_{\substack{1000 \sim 4000 \si{\hertz}}} \left( \bm{v} \right)
    \label{eq:ae-bpf}
\end{eqnarray}
In this way, it is expected that the generated adversarial examples will acquire robustness such that they function even when frequency bands outside the audible range are cut by a speaker or a microphone.

\subsection{Impulse Response}
\label{sec:proposed-impulse}

Impulse response is the reaction obtained when presented with a brief input signal, called an impulse.
Based on the fact that impulse responses can reproduce the reverberation in the captured environment by convolution,
a method of using impulse responses from various environments in the training of a speech recognition model to enhance the robustness to the reverberation has been proposed \cite{DBLP:conf/interspeech/PeddintiCPK15}.
Similarly, we introduced impulse responses to the generation process in order to make the obtained adversarial example robust to reverberations.

In addition, considering the scenario of attacking numerous devices at once via outdoor speakers or radios, we want the obtained adversarial example to work in various environments.
Therefore, in the same manner as \cite{DBLP:conf/icml/AthalyeEIK18}, we take an expectation value over impulse responses recorded in diverse environments.
Here, \eqnref{eq:ae-bpf} is extended like \eqnref{eq:ae-robust}, where the set of collected impulse responses is $\mathcal{H}$ and the convolution using impulse response $h$ is $\mathop{Conv}_{h}$:
\begin{eqnarray}
    & \mathop{\rm argmin}\limits_{\bm{v}} \! \! \! \! \! & \mathbb{E}_{h \sim \mathcal{H}} \! \left[ \mathop{Loss}\limits_{f} \left( \mathop{MFCC} \left( \tilde{\bm{x}} \right) \! , \bm{l} \right) + \epsilon \left \lVert \bm{v} \right \rVert \right] \nonumber \\
    & \, where                            \! \! \! \! \! & \bar{\bm{x}} = \mathop{Conv}\limits_{h} \left( \bm{x} + \mathop{BPF}\limits_{\substack{1000 \sim 4000\si{\hertz}}} \left( \bm{v} \right) \right)
    \label{eq:ae-impulse}
\end{eqnarray}
In this way, it is expected that the generated adversarial examples will acquire robustness such that they are not affected by reverberations produced in the environment in which they are played and recorded.

\subsection{White Gaussian Noise}
\label{sec:proposed-noise}

White Gaussian noise is given by $\mathcal{N} \left( 0, \sigma^2 \right)$ and used for emulating the effect of many random processes that occur in nature.
For example, it is used in the evaluation of speech recognition models to measure their robustness against the background noise \cite{DBLP:conf/interspeech/HansenP98}.
Consequently, we introduce white Gaussian noise in the generation process in order to make the obtained adversarial example robust to background noise.
Here, \eqnref{eq:ae-impulse} is extended as follows:
\begin{eqnarray}
    & \! \! \mathop{\rm argmin}\limits_{\bm{v}} \! \! \! \! \! & \mathbb{E}_{h \sim \mathcal{H}, \bm{w} \sim \mathcal{N} \left( 0, \sigma^2 \right)} \! \left[ \mathop{Loss}\limits_{f} \left( \mathop{MFCC} \left( \tilde{\bm{x}} \right) \! , \bm{l} \right) + \epsilon \left \lVert \bm{v} \right \rVert \right] \nonumber \\
    & where                                     \! \! \! \! \! & \bar{\bm{x}} = \mathop{Conv}\limits_{h} \left( \bm{x} + \mathop{BPF}\limits_{\substack{1000 \sim 4000\si{\hertz}}} \left( \bm{v} \right) \right) + \bm{w}
    \label{eq:ae-noise}
\end{eqnarray}
In this way, it is expected that the generated adversarial examples will acquire robustness such that they are not affected by noise caused by recording equipment and the environment.
Note that the white Gaussian noise should also be added before the convolution for the purpose of emulating thermal noise caused in both the playback and recording devices.
However, we added the noise only after the convolution because doing so makes the optimization easier and \eqnref{eq:ae-noise} was sufficiently robust in the empirical observations.

\section{Evaluation}
\label{sec:evaluation}

In order to confirm the effectiveness of the proposed method, we conducted evaluation experiments.
We played and recorded audio adversarial examples generated by the proposed method and verified whether these adversarial examples are recognized as target phrases.

\subsection{Implementation}
\label{sec:evaluation-implementation}

We implemented \eqnref{eq:ae-noise} using TensorFlow\footnote{Our full implementation is available at \url{https://github.com/hiromu/robust_audio_ae}}.
Since calculating the expected value of the loss is difficult, we instead evaluated the sample approximation of \eqnref{eq:ae-noise} with respect to a fixed number of impulse responses sampled randomly from $\mathcal{H}$.
For optimization, we used Adam \cite{DBLP:journals/corr/KingmaB14} in the same manner as \cite{DBLP:conf/sp/Carlini018}.

\subsection{Settings}
\label{sec:evaluation-settings}

For the input sample $\bm{x}$, we prepared two different audio clips of four seconds cut from \textit{Cello Suite No.~1} by Bach and \textit{To The Sky} by Owl City.
The first clip is the same as the publicly released samples\footnote{\url{https://nicholas.carlini.com/code/audio_adversarial_examples/}} of \cite{DBLP:conf/sp/Carlini018}.
The second clip is the same as the publicly released samples\footnote{\url{https://sites.google.com/view/commandersong/}} of \cite{DBLP:conf/uss/YuanCZLL0ZH0G18}.
The difference between the clips is that the first clip is an instrumental piece and does not include singing voices, whereas singing voices are included in the second song by Owl City.

For the target phrase $\bm{l}$, we prepared three different cases: ``hello world,'' ``open the door\footnote{This phrase is used in \cite{DBLP:conf/uss/YuanCZLL0ZH0G18} to discuss an attack scenario using voice commands.},'' and
``ok google\footnote{This phrase is used as a trigger word of Google Home.}.''
Considering that \cite{DBLP:conf/sp/Carlini018} tested their method with 1,000 phrases that were randomly chosen from a speech dataset, three phrases appear to be insufficient to evaluate the efficiency of our attack.
However, unlike the direct attack as performed by \cite{DBLP:conf/sp/Carlini018}, our evaluation involves a number of playback cycles in the physical world.
This means that our experimental evaluation in the over-the-air setting requires actual time for playing back the generated audio adversarial examples.
For example, our evaluation of a single combination of the input sample and the target phrase requires more than 18 hours in a quiet room without interruption because it involves playing 500 intermediate examples 10 times each with an interval of several seconds.
For this reason, we focused on these three phrases considering the attack scenarios.

For the set of impulse responses $\mathcal{H}$, we collected 615 impulse responses from various databases \cite{DBLP:conf/waspaa/KinoshitaDYNSKM13,DBLP:conf/lrec/NakamuraHANY00,5201259,Wen06evaluationof,harma2001acoustic},
which are constructed primarily for research on dereverberation.

\begin{figure}[tb]
    \begin{center}
        \includegraphics[width=0.48\textwidth]{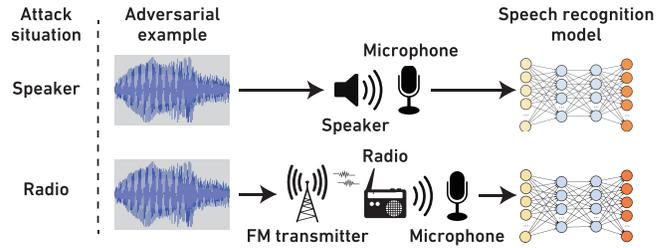}
        \caption{Two attack situations of the evaluation: speaker and ratio. In the first situation, the adversarial examples were played and recorded by a speaker and a microphone. In the second situation, the adversarial examples were broadcasted using an FM radio.}
        \label{fig:situation}
    \end{center}
\end{figure}

For the playback and the recording, we prepared two different attack situations, as shown in \figref{fig:situation}, in order to confirm that the attack using the generated adversarial examples is applicable via a wide range of offensive means.
First, we played and recorded the adversarial examples using a speaker and a microphone (JBL CLIP2 / Sony ECM-PCV80U) in a meeting room with a distance of approximately 0.5 meters.
We also examined whether the generated adversarial examples could attack through the radio using HackRF One\footnote{\url{https://greatscottgadgets.com/hackrf/}}, a Software Defined Radio (SDR) capable of transmission or reception of radio signals.
We broadcasted at 180.0MHz FM and received the signal with a portable radio (Sony ICF-P36) in the same room, while the playback was recorded by a microphone (Sony ECM-PCV80U).
In both cases, we played each adversarial example 10 times to suppress random fluctuation in the physical world and evaluated the recognition results obtained by DeepSpeech.

\subsection{Metrics}
\label{sec:evaluation-metrics}

For the evaluation metrics of the obtained adversarial example, we used the signal-to-noise ratio (SNR) of the perturbation, the success rate of the attack, and the edit distance of the recognition results.
The SNR is given by $10 \log_{10} \frac{P_x}{P_v}$ for the power of the input sample $P_x = \frac{1}{T} \sum_{t=1}^{T} x_t^2$ and the power of perturbation $P_v = \frac{1}{T} \sum_{t=1}^{T} v_t^2$.
In other words, a larger SNR is associated with a smaller perturbation and a smaller likelihood for a human to notice.

The success rate of the attack is the ratio of the number of times that DeepSpeech transcribed the recorded adversarial example as the target phrase among all trials.
The success rate becomes non-zero only when DeepSpeech transcribes adversarial examples as the target phrase perfectly.

Thus, we also introduced the edit distance between the recognition results and the target phrase to confirm the progress of the generation process.
The edit distance reveals the progress more precisely, even when the success rate is 0\%.
Here, the edit distance is defined as the minimum number of procedures required to transform one string into the other by inserting, deleting, and replacing one character.

\subsection{Results}
\label{sec:evaluation-results}

\begin{figure*}[tb]
    \begin{center}
        \includegraphics[width=0.96\textwidth]{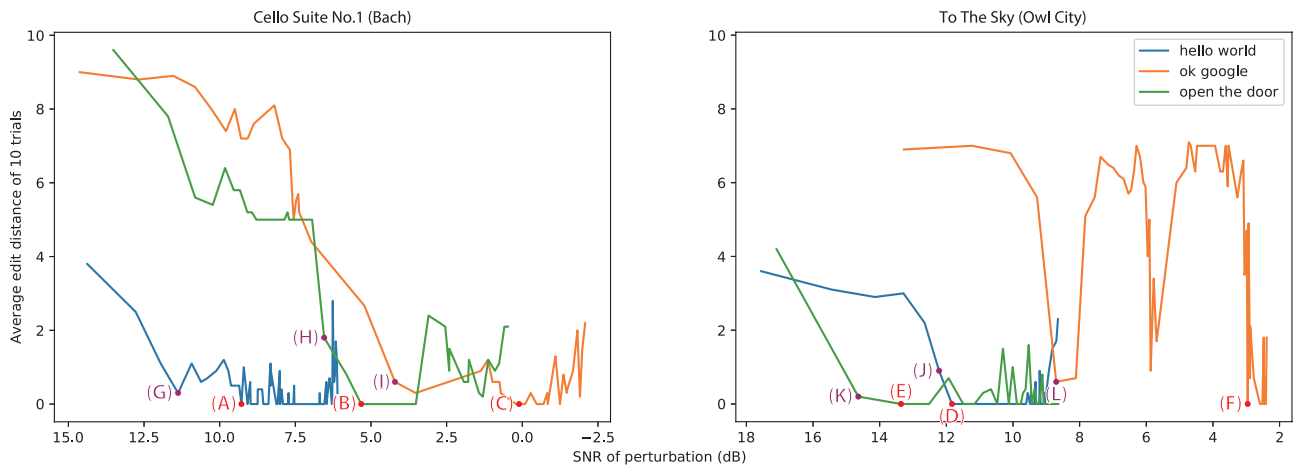}
        \caption{Progress of the generation process of the adversarial examples. As the generation progresses, the SNR and the edit distance in the speaker situation decreases. The detailed results of the highlighted adversarial examples are shown in \tabref{tab:result} and \tabref{tab:result-half}.}
        \label{fig:result}
    \end{center}
\end{figure*}

The progress of the generation process is presented in \figref{fig:result}.
The figure shows that, as the generation progresses, the SNR decreases and the edit distance of the recognition results to the target phrase also decreases.
The detailed results of the generated adversarial examples showed certain levels of the success rate are presented in \tabref{tab:result}.

\begin{table}[tb]
    \centering
    \begin{tabular}{c|c|c|c}
            & Input sample & Target phrase & SNR                  \\ \hline
        (A) & Bach         & hello world   & $9.3\si{\decibel}$  \\ 
        (B) & Bach         & open the door & $5.3\si{\decibel}$  \\ 
        (C) & Bach         & ok google     & $0.2\si{\decibel}$  \\ \hline 
        (D) & Owl City     & hello world   & $11.8\si{\decibel}$ \\ 
        (E) & Owl City     & open the door & $13.4\si{\decibel}$ \\ 
        (F) & Owl City     & ok google     & $2.6\si{\decibel}$  \\ \hline 
    \end{tabular}
    \caption{Details of the generated audio adversarial examples, which showed 100\% success by both the speaker and the radio and having the maximum value of SNR\protect\footnotemark.}
    \label{tab:result}
\end{table}

\footnotetext{\label{sites}These audio files are available at \url{https://yumetaro.info/projects/audio-ae/}.}

\begin{table}[tb]
    \setlength{\tabcolsep}{3pt}
    \centering
    \begin{tabular}{c|c|c|c|c|c|c}
                             & Input                                   & Target                                       & \multirow{2}{*}{SNR}                 & Attack    & Success & Edit  \\
                             & sample                                  & phrase                                       &                                      & situation & rate    & dist. \\ \hline
        \multirow{2}{*}{(G)} & \multirow{2}{*}{Bach}                   & \multirow{2}{*}{\shortstack{hello\\world}}   & \multirow{2}{*}{$11.9\si{\decibel}$} & Speaker   & 60\%    & 1.1   \\ \cline{5-7} 
                             &                                         &                                              &                                      & Radio     & 50\%    & 1.3   \\ \hline
        \multirow{2}{*}{(H)} & \multirow{2}{*}{Bach}                   & \multirow{2}{*}{\shortstack{open\\the door}} & \multirow{2}{*}{$6.6\si{\decibel}$}  & Speaker   & 60\%    & 1.8   \\ \cline{5-7} 
                             &                                         &                                              &                                      & Radio     & 60\%    & 1.8   \\ \hline
        \multirow{2}{*}{(I)} & \multirow{2}{*}{Bach}                   & \multirow{2}{*}{\shortstack{ok\\google}}     & \multirow{2}{*}{$4.2\si{\decibel}$}  & Speaker   & 80\%    & 0.6   \\ \cline{5-7} 
                             &                                         &                                              &                                      & Radio     & 70\%    & 0.9   \\ \hline
        \multirow{2}{*}{(J)} & \multirow{2}{*}{\shortstack{Owl\\City}} & \multirow{2}{*}{\shortstack{hello\\world}}   & \multirow{2}{*}{$12.2\si{\decibel}$} & Speaker   & 70\%    & 0.9   \\ \cline{5-7} 
                             &                                         &                                              &                                      & Radio     & 50\%    & 1.5   \\ \hline
        \multirow{2}{*}{(K)} & \multirow{2}{*}{\shortstack{Owl\\City}} & \multirow{2}{*}{\shortstack{open\\the door}} & \multirow{2}{*}{$14.6\si{\decibel}$} & Speaker   & 90\%    & 0.2   \\ \cline{5-7} 
                             &                                         &                                              &                                      & Radio     & 100\%   & 0.0   \\ \hline
        \multirow{2}{*}{(L)} & \multirow{2}{*}{\shortstack{Owl\\City}} & \multirow{2}{*}{\shortstack{ok\\google}}     & \multirow{2}{*}{$8.7\si{\decibel}$}  & Speaker   & 90\%    & 0.6   \\ \cline{5-7} 
                             &                                         &                                              &                                      & Radio     & 70\%    & 0.9   \\ \hline
    \end{tabular}
    \caption{Details of the generated audio adversarial examples, which showed at least 50\% success by both the speaker and the radio and having the maximum value of SNR\cref{sites}.}
    \label{tab:result-half}
\end{table}

As shown in \tabref{tab:result}, in all combinations of the input sample and the target phrase, the proposed method generated adversarial examples that showed 100\% success by both the speaker and the radio.
On the other hand, the magnitude of the perturbation required to achieve 100\% success differed depending on the input sample and the target phrase.
In the previous method \cite{DBLP:conf/uss/YuanCZLL0ZH0G18} targeted at Kaldi \cite{Povey_ASRU2011} under the over-the-air condition, an SNR of less than 2.0 $\si{\decibel}$ was reported in all cases.
In other words, considering that (D) through (F) of \tabref{tab:result} use the same input sample, the proposed method is able to generate an adversarial example with less perturbation while targeted at a more complex speech recognition model.

\tabref{tab:result-half} showed the adversarial examples having a success rate of at least 50\% by both the speaker and the radio with the maximum value of SNR.
We found that much less perturbation was required to achieve a success rate of 50\%, as compared to \tabref{tab:result}, in all cases.
In other words, the attack using these adversarial examples will succeed once in two attempts and can be a major threat when the attacker allows uncertainty of the attack.

In all cases in \tabref{tab:result} and \tabref{tab:result-half}, the adversarial examples generated with Bach's Cello Suite No.~1 have larger SNR compared to the case of Owl City.
This result supports the discussion of \cite{DBLP:conf/uss/YuanCZLL0ZH0G18}, whereby some phonemes from singing voices in the input sample work together with the injected small perturbations to form the target phrases.
Considering such an effect, we can determine that the result is due to the song by Owl City having more phonemes to help form the target phrases, as compared to Bach's instrumental piece, and requires less perturbation.

We also found that the recognition results of the adversarial examples in \tabref{tab:result-half} changed only slightly between the cases of the speaker and the radio.
This result suggests that the proposed method makes the generated adversarial example robust for FM transmission also.
For example, the addition of white Gaussian noise in the proposed method would also work for the noise caused by FM transmission.
Moreover, as mentioned in \secref{sec:introduction}, one of the major concerns of an audio adversarial example is that it can attack numerous targets simultaneously.
In this respect, the success of the attack through the radio might have a significant impact because such an attack can be made without making victims play an adversarial example actively on their own.

\subsection{Effect of Each Technique}
\label{sec:evaluation-technique}

We then investigated the individual effect of the three techniques on the success of the proposed method.
In detail, we evaluated the effect of the three techniques with changing the combinations in the generation.
Once we obtained an adversarial example that is recognized as the target phrase using the speaker in a similar environment as \secref{sec:evaluation-settings}, we compared its SNR in \tabref{tab:result}.
Here, we used ``hello world'' as the targeted phrase because it is suggested to be relatively easy to generate according to \tabref{tab:result}.

\begin{table}[tb]
    \centering
    \begin{tabular}{ccc|cc}
        \multicolumn{3}{c|}{Used techniques} & \multicolumn{2}{c}{Input sample}                                 \\
        \hline
        Band-pass & Impulse   & Gaussian     & \multirow{2}{*}{Bach}           & Owl                            \\
        filter    & response  & noise        &                                 & City                           \\
        \hline
                  &           &              & --                              & --                             \\
        \ding{51} &           &              & --                              & --                             \\
                  & \ding{51} &              & --                              & --                             \\
                  &           & \ding{51}    & --                              & --                             \\
        \ding{51} & \ding{51} &              & --                              & --                             \\
        \ding{51} &           & \ding{51}    & $-4.2\si{\decibel}$             & $-3.8\si{\decibel}$            \\
                  & \ding{51} & \ding{51}    & --                              & --                             \\
        \ding{51} & \ding{51} & \ding{51}    & $9.3\si{\decibel}$\footnotemark & $11.8\si{\decibel}$\cref{from} \\
        \hline
    \end{tabular}
    \caption{Results of switching in the presence of each technique. Only the case of combining the band-pass filter and white Gaussian noise succeeded to generate, though it requires much more perturbation than the case of \protect\tabref{tab:result}.}
    \label{tab:compare-result}
\end{table}

\footnotetext{\label{from}These values are from \tabref{tab:result}.}

The results are shown in \tabref{tab:compare-result}.
We note that the generation without any of the proposed techniques is equivalent to \cite{DBLP:conf/sp/Carlini018}.
Here, all the cases except the combination of the band-pass filter and white Gaussian noise could not generate adversarial examples that can attack under the over-the-air condition.
In addition, the succeeded combination requires much more perturbation than the case of using all the three techniques.

From these results, it is suggested that we can generate adversarial examples that work over the air without the help of the impulse responses by using white Gaussian noise, whereas the band-pass filter seems to be essential considering the limitation of physical devices.
At the same time, the impulse responses are considered to make adversarial examples robust specifically for reverberations, which results in the reduction of the perturbation, as discussed in \secref{sec:proposed-impulse}.

\section{Listening Experiment}
\label{sec:listening}

In order to consider an attack scenario using the generated adversarial examples, whether humans can notice is important.
If an attacker can make intended phrases to be recognized without it being noticed by humans, then an attack exploiting speech recognition devices will be possible.

For example, \cite{DBLP:conf/uss/YuanCZLL0ZH0G18} conducted listening experiments using Amazon Mechanical Turk (AMT) in the proposal of the attack for Kaldi \cite{Povey_ASRU2011}.
As a result, they reported that only 2.2\% of the participants realized that the lyrics had changed from the original songs used as input samples, whereas approximately 65\% noticed abnormal noises in the generated adversarial examples.

We similarly conducted listening experiments using AMT in order to confirm whether humans notice an attack.

\begin{table}[tb]
    \setlength{\tabcolsep}{5pt}
    \centering
    \begin{tabular}{cccc}
        \hline
        ok google  & turn off  & open the door & happy birthday   \\
        good night & call john & hello world   & airplane mode on \\
        \hline
    \end{tabular}
    \caption{List of choices presented to participants in the listening experiments. We chose simple phrases of lengths similar to those of ``hello world'' or ``open the door,'' concentrating on phrases that are used as voice commands.}
    \label{tab:listen-choice}
\end{table}

\subsection{Settings}
\label{sec:listening-settings}

We used the six generated adversarial examples (A) through (F) of \tabref{tab:result}, which were recognized as target phrases with a success rate of 100\%.
We conducted an online survey separately for each adversarial example with 25 participants.
They listened to the adversarial example three times, and, after each listening, we asked each of the following questions:
\begin{enumerate}
    \item Did you hear anything abnormal? (For affirmative responses, we asked them to write what they felt.)
    \item \textit{(With the disclosure that some voice is included)} did you hear any words? (For affirmative responses, we asked them to write down the words.)
    \item \textit{(With the presentation of eight phrases in \tabref{tab:listen-choice})} which phrase do you believe was included?
\end{enumerate}

\begin{table}[tb]
    \setlength{\tabcolsep}{5pt}
    \centering
    \begin{tabular}{c|c|c|ccc}
            & Hear     & Hear     & \multicolumn{3}{c}{With presentation of}                                     \\ [-0.1em]
            & anything & a target & \multicolumn{3}{c}{the choices listed in \protect\tabref{tab:listen-choice}} \\ \cline{4-6}
            & abnormal & phrase   & Correct & Incorrect & Not sure \\ \hline
        (A) & 36.0\%   & 0.0\%    & 4.0\%   & 28.0\%    & 68.0\%   \\
        (B) & 56.0\%   & 0.0\%    & 4.0\%   & 32.0\%    & 64.0\%   \\
        (C) & 48.0\%   & 0.0\%    & 4.0\%   & 24.0\%    & 72.0\%   \\ \hline
        (D) & 32.0\%   & 0.0\%    & 4.0\%   & 28.0\%    & 68.0\%   \\
        (E) & 44.0\%   & 0.0\%    & 8.0\%   & 16.0\%    & 76.0\%   \\
        (F) & 48.0\%   & 0.0\%    & 0.0\%   & 32.0\%    & 68.0\%   \\ \hline
    \end{tabular}
    \caption{Results of the listening experiments of \protect\tabref{tab:result}. Although a certain number of participants felt abnormal, most of the participants could not hear the target phrases, even when presented with choices.}
    \label{tab:listen-result}
\end{table}

\subsection{Results}
\label{sec:listening-results}

The results are shown in \tabref{tab:listen-result}.
Although a certain number of participants felt abnormal, no one could hear the target phrases in all cases.

In detail, for example, only 32\% of the participants felt abnormal about \tabref{tab:listen-result}(D) and provided comments like, ``It was not very clear,'' ''The music seemed a bit fuzzy,'' and ``It sounded like birds in the background.''
Although \tabref{tab:listen-result}(B) showed the highest rate of the participants feeling abnormal, only comments similar to (D), such as, ``It was like hearing over a bad Skype connection or phone call,'' were provided.
For (D) through (F) of \tabref{tab:listen-result}, which are generated on the song by Owl City, we found more comments related to the sound quality, such as, ``It sounds highly compressed,'' compared to the case of Bach's Cello Suite.
However, an indication of any messages or utterances was not available in all cases.

Furthermore, even when presented with choices for the target phrases, more than half of the participants responded, ``I could not catch anything.''
In particular, no one could choose the correct choice, even though seven participants chose the incorrect choices for \tabref{tab:listen-result}(F).
Moreover, in the other five adversarial examples, only one or two participants chose the target phrase correctly.
Note that these results were obtained under the condition in which we explicitly instructed the participants to listen for the adversarial examples and presented them with choices for the target phrases.
Thus, we believe this result does not deter the attack scenario, which usually seeks a situation that is less likely to be noticed.

Based on the above considerations, we conclude that the generated adversarial examples sound like mere noise and are almost unnoticeable to humans, which can be a real threat.
In addition, the obtained comments suggest directions for future investigation of attack scenarios.
For example, we might able to use birdsong as the input samples or play the samples through a telephone to make adversarial examples more difficult to notice.

\section{Conclusion}
\label{sec:conclusion}

In this research, we proposed a method by which to generate audio adversarial example targeting the state-of-the-art speech recognition model that can attack practically in the physical world.
We were able to generate such robust adversarial examples by introducing a band-pass filter, impulse response, and white Gaussian noise to the generation process in order to simulate the transformations caused by the over-the-air playback.
In the evaluation, we confirmed that the adversarial examples generated by the proposed method can have smaller perturbations than the conventional method, which cannot deal with recurrent networks.
Moreover, the results of listening experiments confirmed that the obtained adversarial examples are almost unnoticeable to humans.
To the best of our knowledge, this is the first approach to successfully generate audio adversarial examples for speech recognition models that use a recurrent network in the physical world.

In the future, we would like to examine a detailed attack scenario and possible defense methods regarding the generated audio adversarial examples.
We would also like to consider the possibility of realizing a robust speech recognition model using adversarial training, as discussed for the image classification \cite{DBLP:journals/corr/GoodfellowSS14}.

\section*{Acknowledgments}

This work was partly supported by KAKENHI (Grants-in-Aid for scientific research) Grant Numbers JP19H04164 and JP18H04099.

\bibliographystyle{named}
\bibliography{paper}

\end{document}